# A Unified Multi task Learning Architecture for Hate Detection Leveraging User-based Information

**Prashant Kapil**
Department of CSE
Indian Institute of Technology Patna
`prashant.pcs17@iitp.ac.in`

**Asif Ekbal**
Department of CSE
Indian Institute of Technology Patna
`asif@iitp.ac.in`

## Abstract

Hate speech, offensive language, aggression, racism, sexism, and other abusive language is a common phenomenon in social media. There is a need for Artificial Intelligence (AI) based intervention which can filter hate content at scale. Most existing hate speech detection solutions have utilized the features by treating each post as an isolated input instance for the classification. This paper addresses this issue by introducing a unique model that improves hate speech identification for the English language by utilising intra-user and inter-user-based information. The experiment is conducted over single-task learning (STL) and multi-task learning (MTL) paradigms that use deep neural networks, such as convolution neural network (CNN), gated recurrent unit (GRU), bidirectional encoder representations from the transformer (BERT), and A Lite BERT (ALBERT). We use three benchmark datasets and conclude that combining certain user features with textual features gives significant improvements in macro-F1 and weighted-F1.

## 1 Introduction

The rise in social media platforms enables a user to express their opinions, inner thoughts or emotions through social networks like Twitter and Facebook. Social media's broad use has led to an increase in the number of nasty posts being shared online. The moderators are burdened by the quantity of content and the psychological stress of reviewing harmful posts. To combat this, an automated system is required to detect hate speech and assist the moderators in rejecting or approving the post. According to the statistics, passive or manual methods (like flagging) of preventing the spread of hate speech are neither efficient nor readily scaleable (Pavlopoulos et al., 2017). Hate speech is an umbrella term that covers instances of abusive, offensive, racist, sexist, aggressive, etc. This kind of behaviour is detrimental from a social and business perspective. The detection of hate speech is a challenging task due to limited text information, grammar and syntactic flaws (Founta et al., 2018). A user tending to post hate messages might do it again. Having some background information about the user of a post may be very informative. The hypothesis is that by infusing user's historical tweets along with the information present in the tweet, the proposed system will be having good accuracy. The key attributes of our current work are as follows:

1. **Model**: The experiment is performed over CNN, GRU, BERT and ALBERT in single-task learning (STL) and multi-task learning (MTL) paradigms. The MTL framework is also augmented with intra-user, inter-user, and tweet based features.

2. **Data**: Three publicly available data sets with the tweet ID information is leveraged.

3. **Result and Error Analysis**: The results are analyzed along with the qualitative analysis.

## 2 Related Work

Deep learning-based techniques are now being used to tackle the problem of supervised learning to address hate speech. Most of the existing research merely uses tweet information for classification. However, some of the works have also made use of inter-user and intra-user data from historical posts made by the user. (Qian et al., 2018) leverages inter-user and intra-user representation learning for hate speech detection on Twitter. The user's historical posts are incorporated as intra-user representations. The semantically similar tweets posted by all the other users are utilized as inter-user features. (Pitsilis et al., 2018) proposed a detection scheme that is an ensemble of recurrent neural network(RNN) classifiers, incorporating various features associated with user-related information, such as users' tendency towards racism or sexism. (Founta et al., 2019) proposed a deep learning architecture that utilizes a wide variety of metadata such as tweet-based, user-based, and network-based features. (Chatzakou et al., 2017) investigated 30 features from three attributes such as text, user, and network-based attributes to study the properties of bullies and aggressors. (Rajadesingan et al., 2015) derived 10 features grouped into text-based features, emotion-based features, familiarity-based features, contrast-based features, and complexity-based features to solve the sarcasm detection. (Waseem and Hovy, 2016) leveraged the gender and demographic information as a feature to improve the performance. (Unsvåg and Gambäck, 2018) investigates the potential effects of users' features on hate speech detection. The different user features were incorporated, such as gender, network (number of followers and friends), activity (number of statuses and favourites), and profile information (geo-enabled, default profile, default image, and number of public lists). (Chaudhry and Lease, 2022) investigate profiling users by their past utterances as an informative prior to better predict whether new utterances constitute hate speech. (Pougué-Biyong et al., 2023) (Vosoughi et al., 2016) (Mishra and Diesner, 2018)

## 3 Data

Three publicly accessible data with the tweet ID are used in the experiment. The past tweets are collected using the Twitter API. The data is explained here.

**D1 (Founta et al., 2018)**: It includes the tweet id and 99,996 tweets classified with the labels, "hate," "abusive," "spam," and "neutral." The tweets were selected based on the sentiment analysis, showing strong negative polarity ($< -0.7$) and contain at least one offensive word. Using Twitter API, the meta tweet is crawled consisting of user meta information. We were able to collect 17,999 meta-tweets because the majority of the accounts were either suspended or the tweet was removed.

**D2 (Zampieri et al., 2020)**: provided semi-supervised offensive language identification dataset (SOLID) which contains 9,089,140 tweets id with the label obtained in a semi-supervised manner using democratic co-training with OLID (Zampieri et al., 2019b) as the seed data. The tweets with a probability value $\geq 0.85$ for the offensive and non-offensive posts were taken. In addition to 'gun control', and 'to:BreitbartNews' used during the trial annotation, four new political keywords were used to collect tweets for the full dataset: 'MAGA', 'antifa', 'conservatives', and 'liberals'.

**D3 (Waseem and Hovy, 2016)**: The 16,000 tweets were collected using keywords and phrases such as "MKR", "asian drive", "feminazi", "immigrant", "nigger", "sjw", "WomenAgainstFeminism", "blameoneonotall", "islam terrorism", "notallmen","victimcard", "victim card", "arab terror", "gamergate", "jsil", "racecard", "race card" and divided into three categories: neural, sexism, and racism. As there were only 9 meta tweets for the racism class, only sexism and neural tweet ID were taken into account to collect the historic posts.

The data statistics is shown in Table 1.

| Dataset | labels and count | Avg.His.post(30) | #Posts |
|---------|------------------|------------------|--------|
| $D_1$   | Spam: 3117       | 3.5              |        |
|         | Abusive: 2584    | 5.8              | 17999  |
|         | Hateful: 764     | 7                |        |
|         | Neutral: 11534   | 2.5              |        |
| $D_2$   | Offensive: 12439 | 4.5              | 34565  |
|         | NOT: 22126       | 1.8              |        |
| $D_3$   | Sexism: 994      | 7                | 3308   |
|         | Neutral: 2314    | 2.25             |        |

Table 1: **Data Statistics**

## 4 Methodology

The proposed model is compared with several strong baselines:
**CNN**: It is proposed by (Kim, 2014) and consists of five main layers: input layer, embedding layer, convolution, pooling, and fully connected layer.

**GRU**: This is proposed by (Cho et al., 2014). The features are extracted by using two gating mechanisms, *viz.* reset and update.

**BERT** (Devlin et al., 2018): It extracts the features by applying the bidirectional training of the transformer to learn the context of a word based on all of its surrounding leveraging masked language modelling (MLM) and next sentence prediction (NSP).

**ALBERT**: It is a lite BERT for self-supervised learning of language representations to learn the features by MLM and sentence ordering prediction

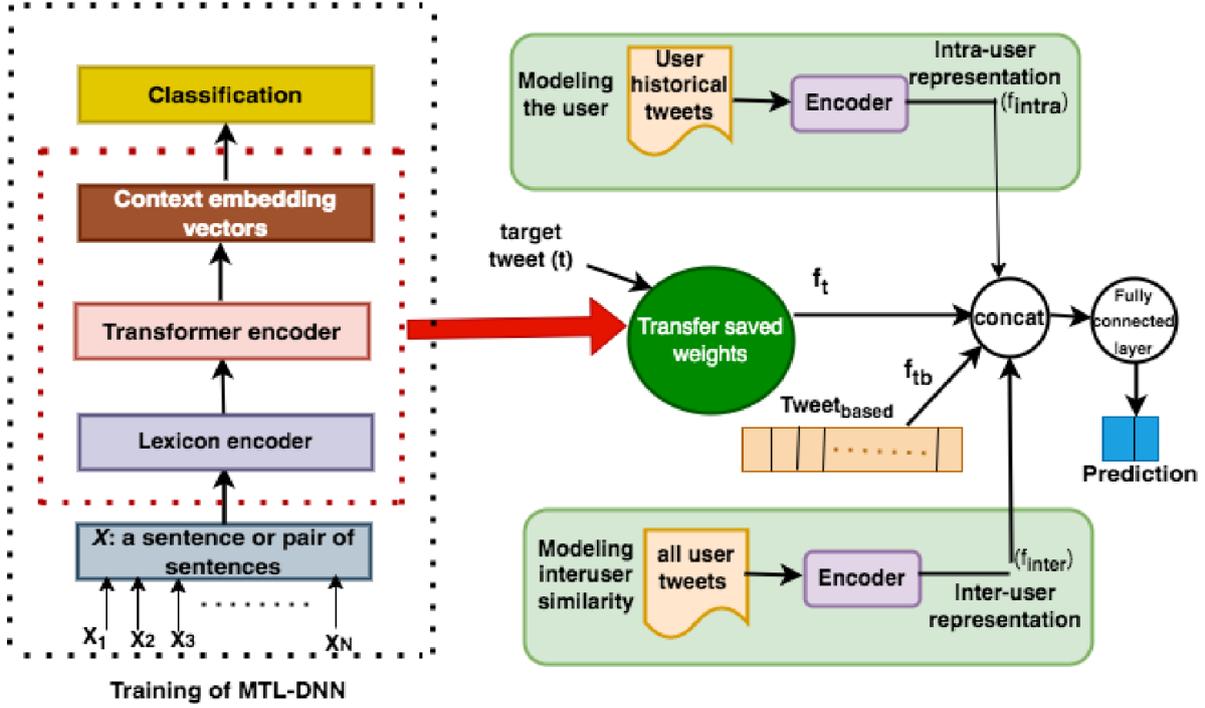

Figure 1: **Target tweet features concatenated with inter-user, intra-user and tweet based representation**

(SOP).

The aforementioned four models were trained in a STL paradigm. The BERT model is selected to train the three datasets jointly in the MTL setup. It aims at solving more than one classification problem simultaneously. The end-to-end deep multi-task learning has been recently employed in solving various problems of natural language processing (NLP), such as sentiment, emotion, and hate prediction.
(Zhang and Yang, 2017) defined MTL as follows:
*Definition 1 : Multi-Task learning:* Given $m$ learning tasks

$$\{T_i\}_{i=1}^{m} \qquad (1)$$

where all the tasks or subset of them are related, multi-task learning aims to help improve the learning of a model for classification task $T_i$ by using the knowledge in some or all of the $m$ tasks.
The MTL model is evaluated on three different datasets related to hate speech classification, sexism detection, and offensive language detection. It is explained as follows:
**Multitask learning (MTL-DNN):** It is proposed by (Liu et al., 2019) for learning representation across multiple natural language understanding tasks. The lower layers are shared across each task while the top layers are task-specific. The input $X$ is first represented as a sequence of embedding vectors for each word. This is followed by the transformer encoder capturing the contextual information for each word and generating the shared contextual embedding vectors. In the last layer, additional task-specific layers generate task-specific representations which are then passed on to the final classification. The weight matrix associated with the shared layer, marked in red in Figure 1, is transferred to the new model to obtain the shared features. Along with the shared features, meta features such as tweet based, intra-user, and inter-user features were also considered.

These metadata information plays an important role in improving the performance and described as follows:
**Tweet based(TB)($f_{tb}$):** The features such as the number of hashtags, number of emoticons, number of uppercase and lowercase letters, number of positive and negative words, and sentiment score by (Hutto and Gilbert, 2014) were considered. Each tweet is lowercased and stemmed using the Porter stemmer followed by creating unigram, bigram, and trigram features each weighted by its TF-IDF.
**Intra-User tweets($f_{intra}$):** Given a tweet ID, the user's historical post is collected by Twitter API. Given a tweet $t$, we collected $m$ tweets posted by this user i.e. $Z_t = \{z_1, z_2,....z_m\}$. These intra-user tweets are passed through BERT to get $f_{intra}$. The value of $m$ is 200 for each user and used in the batches of 50, 100, and 200 during training.

| Model | D1 | | D2 | | D3 | |
|---|---|---|---|---|---|---|
| | Macro(%) | Weighted(%) | Macro(%) | Weighted(%) | Macro(%) | Weighted(%) |
| Single Task Learning (STL) | | | | | | |
| CNN | 64.22 | 78.65 | 92.71 | 92.73 | 82.61 | 84.23 |
| GRU | 60.92 | 74.26 | 90.77 | 92.73 | 85.02 | 86.47 |
| BERT | 67.46 | 81.36 | 95.12 | 95.72 | 86.72 | 87.91 |
| ALBERT | 59.39 | 75.44 | 91.22 | 91.42 | 79.91 | 80.88 |
| Multi Task Learning (MTL) | | | | | | |
| CNN | 69.71 | 80.54 | 94.96 | 95.12 | 86.88 | 89.12 |
| GRU | 66.92 | 78.87 | 92.93 | 93.24 | 88.24 | 89.73 |
| BERT | 72.58 | 84.47 | 97.90 | 98.12 | 89.34 | 90.83 |
| ALBERT | 67.24 | 81.66 | 92.27 | 92.98 | 84.48 | 86.70 |
| Proposed | | | | | | |
| MTL-BERT+Intra | 73.08 | 85.12 | 98.12 | 98.23 | 90.46 | 91.92 |
| MTL-BERT+Inter | 73.25 | 85.78 | 98.28 | 98.34 | 90.81 | 92.47 |
| MTL-BERT+Intra+Inter+TB | **73.78** | **86.02** | 98.41 | 98.67 | 90.99 | 93.15 |

Table 2: **Evaluation Results on $D_1$ (Founta et al., 2018), $D_2$ (Zampieri et al., 2020), and $D_3$ (Waseem and Hovy, 2016)**

**InterUser tweets($f_{inter}$)**: The similarity between users on Twitter is calculated by the formula proposed in (AlMahmoud and AlKhalifa, 2018) which uses seven features. These are following and followers, mention, retweet, favourite, common hashtag, common interests, and profile similarity. We selected three similar users and included their tweets to represent inter-user information. The semantically similar tweets to the target are used to add features to the target tweet. All the similar tweets were passed to the encoder to generate $f_{inter}$.

**Proposed Approach** As shown in Figure 1 the trained weight parameters shown in red obtained from MTL is transferred to get the target tweet feature $t$. The other three features, namely $f_{intra}$, $f_{inter}$, and $f_{tb}$ were concatenated with $t$ and passed on to fully connected layer with softmax activation function to obtain the final prediction.

## 5 Experiment setup

The experiments were performed using a 5-fold cross-validation approach. The 4-fold training set is split into 15% validation and 85% training while the last fold is treated as the test set to evaluate the model. All the deep learning models were implemented using Keras (Chollet et al., 2015)with Tensorflow (Abadi et al., 2016) as the backend. The number of filters used in CNN is 100, and the kernel width ranges from 1 to 4. For the GRU, the number of hidden nodes is set to 100. The concatenation of word2vec embedding (Mikolov et al., 2013) and Fasttext (Bojanowski et al., 2017) is used for CNN and GRU in STL and MTL. Categorical cross-entropy is used as a loss function, and Adam (Kingma and Ba, 2014) optimizer is used for optimizing the network. The learning rate of 2e-5 is used for all the models. The batch size of 32 is used to train the shared encoder and an epoch of 3 is found to be optimal. The value for bias is randomly initialized to all zeros, the relu activation function is employed at the intermediate layer, and softmax is utilized at the last dense layer. To achieve stability in the results produced, every classifier is run for 5 times and the output values were aggregated.

## 6 Results and Analysis

Table 2, show the experimental results obtained for three datasets in STL and MTL. The macro-F1 and weighted-F1 are used as performance metrics. The BERT-based multi-task learning outperforms the single task learning by reducing false positives. The inclusion of intra-user and inter-user features helped in improving the overall performance for all the three datasets. The model significantly improves upon adding all the features. The high performance observed in D2 is attributed to the fact that the instances in SOLID (Zampieri et al., 2020) were generated by an ensemble of four models. Hence the model is overfitting with the correct information in the case of D2.

**Error Analysis**: The qualitative analysis is done by providing one case for the false positive (Non-Hate → Hate) and false negative (Hate → Non-Hate).

**False Positive**
(i) *Obfuscation*
Tweet: *@user you're a b**** that would believe anything.*

**Explanation**: The user with the tendency of hiding the slang word in the target tweet and historical posts makes it difficult for the model to predict correctly.
**False Negative**
(i) *Sarcasm*
*Artists are here to disturb the peace-@user.*
**Explanation**: The user with the tendency of writing sarcasm, satire, or metaphor does it frequently resulting in confusing the model to misclassify a post.
**Classification of historical posts** The existing English data (Davidson et al., 2017), (Waseem and Hovy, 2016), (Kumar et al., 2018), (Zampieri et al., 2019a), (Golbeck et al., 2017), (de Gibert et al., 2018), (Basile et al., 2019), (Mandl et al., 2019), (Bhattacharya et al., 2020), (Modha et al., 2020), (Founta et al., 2018) summed upto 230K is fine-tuned over BERT (Devlin et al., 2018) to get the prediction for the 50 historical posts with respect to each user. The average hateful posts with respect to each class is shown in Table 1. It can be seen that the hateful, offensive, and sexism users had 7, 4.5, and 7.8 number of hateful posts out of 50. The abusive and spam users have slightly lower value of 3.5 and 5.8 hateful posts. The Neutral class in all the users got the lowest value of 2.5, 1.8, and 2.25 hateful posts out of 50 historical posts.

## 7 Conclusion and Future work

In this work, we propose a novel method for hate speech detection by experimenting with CNN, GRU, BERT, and ALBERT in STL and MTL fashion. The augmentation of inter-user, intra-user, and other tweet-based features outperforms the sota approaches in terms of macro-F1 and weighted-F1. Furthermore, intra-user and inter-user representation learning can be generalized to other text classification tasks, where either user history or a large collection of unlabeled data is available. In addition to detecting potential hate speech, this method can also help study the dark triad personality (narcissism, Machiavellianism, and psychopathy).

## Acknowledgement

The Authors gratefully acknowledge the project "HELIOS - Hate, Hyperpartisan, and Hyperpluralism Elicitation and Observer System", sponsored by Wipro Ltd. Prashant Kapil acknowledges the University Grant Commission (UGC) of the Government of India for UGC NET-JRF/SRF fellowship.